\def\year{2022}\relax
\documentclass[letterpaper]{article} 
\usepackage{aaai22}  
\usepackage{times}  
\usepackage{helvet}  
\usepackage{courier}  
\usepackage[hyphens]{url}  
\usepackage{graphicx} 
\usepackage{natbib}  
\usepackage{caption} 
\DeclareCaptionStyle{ruled}{labelfont=normalfont,labelsep=colon,strut=off} 
\usepackage [english]{babel}
\usepackage [autostyle, english = american]{csquotes}
\MakeOuterQuote{"}
\usepackage{algorithm}
\usepackage{algorithmic}

\usepackage{newfloat}
\usepackage{listings}
\floatstyle{ruled}
\newfloat{listing}{tb}{lst}{}
\floatname{listing}{Listing}
\title{ Federated Learning: Balancing the Thin Line Between Data Intelligence and  Privacy\\
}

\author{
    Sherin Mary Mathews\textsuperscript{\rm *}
      , Samuel A. Assefa \textsuperscript{\rm *}
}
\affiliations{
    \textsuperscript{\rm *} U.S. Bank AI Innovation \\

   sherin.mathews@usbank.com, samuel.assefa@usbank.com
}

\usepackage{bibentry}

\begin{document}

\maketitle
\vspace{-0.2cm}
\begin{abstract}
Federated learning holds great promise in learning from fragmented sensitive data and has revolutionized how machine learning models are trained. This article provides a systematic overview and  detailed taxonomy of federated learning. We investigate the existing security challenges in federated learning and provide a comprehensive overview of established defense techniques for data poisoning, inference attacks, and model poisoning attacks. The work also presents an overview of current training challenges for federated learning, focusing on handling non-i.i.d. data, high dimensionality issues, heterogeneous architecture, and discusses several solutions for the associated challenges. Finally, we discuss the remaining challenges in managing federated learning training and suggest focused research directions to address the open questions. Potential candidate areas for federated learning, including  IoT ecosystem, healthcare applications, are discussed with a particular focus on banking and financial domains.  
\end{abstract}

\section{ 1. Landscape of  Federated Learning }
\vspace{0.05cm}
\subsection{1.1 Need for Federated Learning}

Data has always been of significant priority for businesses of all sizes, especially in Financial Services. With the advancement of technology, companies tend to capture customer data from many sources, such as tracking customer's activities and appending other data sources to proprietary sources to enhance their ability to contextualize data and draw new insights. However, preserving customer data privacy is vital considering the sensitive nature of customer data. \cite{hahn2018security,yang2019federated}.\\

Several factors have driven the global consensus to focus on \textbf{preserving data privacy and security} of individuals, businesses, and societies. As a result, governments are strengthening data security and privacy protection measures. For example, General Data Protection Regulations \textbf{(GDPR)}  aims to protect the user's privacy by requiring operants to distinctly obtain user consent and adhere to sufficient data privacy requirements \cite{khan2019data}. The establishment of these laws and regulations poses new challenges to the traditional data processing mode of Artificial Intelligence (AI) to varying degrees. With frequent incidents of personal data breaches and individual and institutional data rights not being equal, there is a need for strict data privacy regulations. As traditional machine learning exposes more of its drawbacks, finding new, secure, and effective ways to collect as well as learn from data becomes crucial, thereby opening new research avenues and methods to build \textbf{personalized models without violating user privacy}. \\

In addition to the privacy-preserving dilemma, there is an additional dilemma related to  \textbf{``data isolated islands''} \cite{li2020preserving}. Data is foundational for building AI models, and data islands lead to data being stored, maintained, and isolated in different organizations \cite{li2019survey}. Integrating the data scattered in various organizations is challenging and could introduce a considerable cost. These challenges are particularly prevalent in the Financial Services industry, where data is often stored and processed in a highly segregated manner due to regulatory and privacy concerns. The following sections discuss the two key challenges and why federated learning is suited for their resolution. \\

\vspace{-0.2cm}
\subsection{1.2 Definition of Federated Learning}
\vspace{-0.05cm}
Federated learning is a distributed machine learning architecture that solves the dilemma of learning across \textbf{data silos} by enabling the training over multiple decentralized data stores. The federated learning setting also generates more robust models without sharing data, leading to \textbf{privacy-preserved solutions} with higher security and tighter access privileges for data \cite{konevcny2015federated,konevcny2016federated}. Traditional machine learning frameworks primarily use the centralized method to process the data using centralized collection, unified processing, cleaning, and training models, which requires the training data to be located in the same server. To this end, federated learning adheres to two important ideas: \textbf{local computing and model transmission}, which reduces systematic privacy risks and costs from traditional centralized machine learning methods. The inherent focus on decentralized training is aimed at ensuring the data privacy of each device \cite{zhang2021survey}.\\

Federated learning brings the model and code to the data rather than taking the data to where the model resides as is currently the case in most learning approaches, i.e., the data remains locked on a server or edge device while only the algorithm travels between the servers \cite{xu2021federated}. Federated learning is known as \textbf{collaborative learning}, where algorithms get trained across multiple devices or servers with decentralized data samples without exchanging the actual data. This approach is radically different from other more established techniques, such as getting the data samples uploaded to servers or having data in some form of distributed infrastructure. 

\vspace{-0.1cm}
\section{2. Taxonomy and System Architecture of Federated Learning}
\vspace{0.05cm}
As data privacy increasingly becomes a critical societal concern, federated learning has been a crucial research topic in enabling the collaborative training of machine learning models among different organizations under privacy restrictions. As researchers try to support more machine learning models with different privacy-preserving approaches, there is a need to develop systems and infrastructures to reduce the complexity and effort of various federated learning algorithms. This section conducts a review of federated learning architectures \cite{li2019survey} and analyzes the system components to understand the critical system design components and guide future research. Compared to other federated learning reviews focused on general communication architecture \cite{yang2019federated}, platforms \cite{aledhari2020federated}, and protocols \cite{lim2020federated,lin2020ensemble}, this paper mainly provides an overview of the federated learning paradigm, focusing primarily on machine learning and security criteria. Specifically, we provide a  taxonomy for federated learning systems according to the following four aspects: \textbf{\textit{data distribution, privacy mechanisms, federation scale, and open source frameworks}}. 


\subsection{ 2.1. Data Distribution}
\vspace{-0.05cm}
While considering how the training and inference data are distributed, existing federated learning approaches can be classified into \textit{horizontal federated learning}, \textit{vertical federated learning, and federated transfer learning}.\\

\vspace{-0.1cm}

\textbf{Horizontal federated learning} is a federated learning approach in which datasets on the devices share the same attributes in different instances \cite{yang2019federated}. In this category of federated learning, users have similar attributes in terms of domain usage style and derived statistical information. An example would be a machine learning model to predict the probability of possible occurrence of cancer cells or a machine learning model for next word prediction and keyword spotting \cite{leroy2019federated}. Federated learning allows the secure sharing of user-sensitive data (e.g. date of birth, account number, medical images) through aggregated updates from each client.\\

\vspace{-0.2cm}
\textbf{Vertical Federated learning} is applicable where shared data between unrelated domains are used to train the global model. Clients utilizing this approach have a transitional resource to provide encryption logic to guarantee that only the common data statistics are shared between separate organizations \cite{yan2016survey,cheng2021secureboost}. A real-time use-case would be a scenario where a bank's marketing team of a credit card division would like to improve their model by learning the most purchased items from external online shopping domains. User information from the bank and details from separate shopping sites are shared to train the model with the intermediate encryption logic, ensuring a restricted and secure exchange of only derived statistics. With this liaising of information exchange, banking domains can serve customers better with relevant offers, and online shopping domains can revise their points allocation for customers using credit card transactional data. \\

\vspace{-0.2cm}
\textbf{Federated Transfer learning} utilizes the classic machine learning-based transfer learning \cite{pan2009survey} technique to train a new requirement on a pre-trained framework that has been already trained on a similar dataset to solve an entirely different problem \cite{liu2020secure}. Training on a pre-trained model gives an advantage compared to using a fresh model built from scratch. A real-time example would be transferring a global model to a personalized user and adjusting the model to provide a customized model on a specific user's wearable device. Similar to vertical federated learning, participants can benefit from larger datasets and well-trained machine learning model statistics to serve their unique requirements using a federated transfer learning approach.
\subsection{2.2. Privacy Mechanisms}
\vspace{-0.05cm}
Privacy mechanisms play a key role in federated learning and offer another way to group algorithms. The two state-of-the-art privacy mechanisms for federated learning-based data protection include {\textit{differential privacy and cryptographic methods}}.\\

\vspace{-0.2cm}
\textbf{Differential Privacy} prevents the federated learning server from identifying the owner of a local update and ensuring that a single record does not influence the output of a function \cite{dwork2006our}. Differential Privacy adds a certain degree of noise in the original local update while furnishing theoretical guarantees on the model quality and protection against the inference attack on the model \cite{cheng2021secureboost,truex2019hybrid}. However, due to the injected noises in the learning process, such systems tend to produce less accurate models. \\

\vspace{-0.2cm}
\textbf{Cryptographic methods} include homomorphic encryption and secure multi-party computation (SMC) \cite{bonawitz2017practical,chai2020secure,fontaine2007survey}. The parties need to encrypt their messages before sending and decrypt the encrypted output leading to high computation overhead. The user privacy of these federated learning systems is usually well protected as the cryptographic method guarantees that all the parties cannot learn anything except the output \cite{hardy2017private}.

\subsection{ 2.3. Scale of Federated Learning }
\vspace{-0.05cm}
The scale of federation is another critical factor in designing effective algorithms. Based on the scale of data and the number of client nodes, federated learning can be labelled as \textit{ Cross-device federated learning} or \textit{Cross-silo federated learning}. 

\vspace{-0.2cm}
\subsection*{Cross-device Federated Learning}
\vspace{-0.05cm}
Cross-device federated learning has many clients in an analogous domain with similar interests. This type is an excellent fit for IoT or mobile applications use-cases \cite{yu2020sustainable}. Due to the significant number of clients, tracking and maintaining transaction history logs is not easy. Most clients connect using unreliable networks where participation in training rounds happens randomly. Similar to data partitioning in horizontal federated learning, resource allocation strategies like client selection and device scheduling are used to make updates.\\

\vspace{-0.2cm}
\subsection{Cross-silo Federated Learning} 
\vspace{-0.05cm}
Clients are of small-scale numbers ranging from 2 to 100 indexed devices and are almost always available for training rounds. Cross-silo federated learning is more flexible than cross-device federated learning \cite{zhang2020batchcrypt}. It is used in scenarios within organizations or within groups of organizations to train the machine learning model with their confidential data. Training data can be horizontal or vertical with vertical learning methods resulting in significant communication bottlenecks and computation issues. Similar to vertical federated learning and federated transfer learning implementations, the inference information is restricted using the homomorphic encryption technique. The batch encryption technique reduces computation and communication costs \cite{zhang2020batchcrypt}.

\subsection{ 2.4 Open-source Federated Learning frameworks}
PySyft, FATE, and Tensorflow Federated are currently a few open-source frameworks for researchers and developers to explore federated learning solutions.  \textbf{PySyft} is written in Python on top of the PyTorch framework and provides a virtual hook for connecting to clients through a websocket port \cite{ryffel2018generic}.   \textbf{FATE} framework provides production-ready APIs with Kubernetes integration to implement federated learning in horizontal, vertical, and transfer learning modes \cite{Fed)}. \textbf{Tensorflow Federated} \cite{Github} includes integration with Google Kubernetes Engine (GKE)  or a Kubernetes cluster for orchestrating interaction with clients and the central server for federated learning. Google's TensorFlow Federated -TFF5 \cite{Github} is one of the first attempts in the community to bring federated learning to practical reality, and Gboard enables android mobile users to predict the next word while using the local mobile phone keyboard.\\

\vspace{-0.15cm}
\section{3. An empirical study on current challenges in Federated Learning}
\vspace{0.05cm}
The topic of federated learning is still in its infancy and will continue to be an active area of research for the foreseeable future. As federated learning evolves, so will the attack mechanisms, and hence it is essential to provide a broad overview of current challenges on federated learning. In this section, we outlined the different challenges and potential vulnerabilities to improve the robustness of federated learning systems. Our ultimate goal is to investigate these areas, promote research collaboration, and develop general-purpose defense mechanisms robust against various attack modalities without degrading model performance.\\

While the applications are many, several challenges are associated with federated learning. These challenges can be broadly classified into two categories: \textbf{security-related challenges} and \textbf{ training challenges}. 

\vspace{-0.1cm}
\subsection{3.1 Security challenges}
\vspace{-0.05cm}
Security-related challenges include the privacy and security threats that arise due to the presence of an adversary who gains access to a user device and installs a malicious client that gains access to the black-box algorithm. Security attacks can be induced mainly by malicious actors in the learning process, and they can be either targeted or non-targeted. In targeted attacks, the adversary wants to influence the prediction on specific tasks, while in non-targeted attacks, the adversary's motivation is to compromise the accuracy of the global model. We will concentrate on model poisoning attacks, data poisoning attacks, and inference attacks within the security vulnerabilities. 
\vspace{-0.1cm}
\subsection{(A) Data Poisoning Attacks}
\vspace{-0.05cm}
In a data poisoning attack, an attacker poisons the training data for a certain number of participating devices during the learning process resulting in compromising the accuracy of the global model. Here, the attacker poisons the data by directly injecting poisoned data to the targeted device or other connected devices and hence is one of the most commonly used attack techniques against machine learning models \cite{chen2017targeted,shafahi2018poison}. Data poisoning attacks also leverage gradient descent to generate adversarial training examples. \\

A label-flipping attack and backdoor attack fall under data poisoning attacks.  In a \textbf{label-flipping attack}, an adversary can alter its local data by flipping the labels of training instances of source class to the target class while keeping the training data features intact, resulting in substantial drops in global model's accuracy \cite{fung2018mitigating,biggio2012poisoning}. In a \textbf{backdoor poisoning attack}, an  adversary inserts backdoored inputs into local data to tweak individual features  that are then transferred into the global model \cite{gu2017badnets}.\\

 Mitigation approaches can make use of PCA-based clustering and techniques such as the FoolsGold framework to defend against data poisoning attacks \cite{fung2018mitigating}. Clustering approaches check model updates at the aggregator and then cluster them into two groups using dimensionality reduction techniques. Clusters identified with less than n/2 clients are placed into suspicious clusters of malicious clients. The FoolsGold scheme proposes limiting potentially malicious client's contributions with similar model updates to the global model by reducing their learning rates. It shows promising results when the training data is non-i.i.d. but fails when the training data is i.i.d. as it incorrectly penalizes honest clients with similar data distributions resulting in substantial drops in test accuracy. Another way to defend against backdoor attack genre is to identify the participants based on their model updates before model averaging in each round of learning.
\vspace{-0.1cm}
\subsection*{(B) Model Poisoning Attacks}
A formidable challenge in federated learning is the possibility of an adversary initiating an attack to poison the local client's models instead of the local data. The attacker compromises some of the local devices modifying its local model parameters, thus shifting the model's boundary, introducing global model errors, and affecting its accuracy. For example, the attacker can introduce a stealthy backdoor functionality into the global model and compromise one or several participants. It trains a model on the backdoor data using a  constrain-and-scale technique, submits the resulting model, and replaces the global model with the attacker's backdoored model. \\

The most common model poisoning defense measures combine secure aggregation, anomaly detection, and participant-level differential privacy. Using secure aggregation acts as a robust defense mechanism as the individual updates from each participant are invisible to the aggregator. Integrating these combinations of solutions into an automatic, predictable model helps prevent poisoning attacks \cite{bhagoji2019analyzing,bagdasaryan2020backdoor}. Additional defense mechanisms against model poisoning attacks include rejections based on error rate, loss function, or a combination of both. In error rate-based rejections, the framework rejects the models with a significant effect on the error rate of the global model. In loss function-based rejections, the models with a significant impact on the loss function of the global model will be rejected. 
\vspace{-0.1cm}
\subsection*{(C) Inference Attacks}

Despite the substantial privacy promise of federated learning, inference attacks have demonstrated that it is possible to infer sensitive personal information from training data used in model updates during the learning process in some scenarios. In inference attacks, an attacker can infer sensitive information to which no access is granted by querying the model several times or using prevailing common knowledge. A commonly used method to mitigate inference attacks is to utilize differential privacy that provides efficient and statistical guarantees against learning for an adversary \cite{su2018securing}. Noise is added to the data to obscure sensitive items so that the other party cannot distinguish the individual's information, making it impossible to restore the original data, thereby rendering inference attacks ineffective\cite{shokri2017membership}. \\

Other defense techniques include calibrated domain-specific data augmentation, in which the distinctiveness of model updates is decremented using calibrated domain-specific data augmentation. Additionally,  running the framework in a trusted execution environment and secure computation are good defense techniques to counteract inference attacks. Trusted Execution Environment (TEE) presents a secure platform for running the federated learning process with low computational overhead but is suitable only for CPU devices. The most commonly used methods used in Secure Computation include \textbf{\textit{Homomorphic Encryption and  Secure Multiparty Computation (SMC)}}. In homomorphic encryption, computations are executed on encrypted inputs without decrypting the data. In SMC, two or more parties concede to run the inputs provided by the clients and expose the outputs only to a subset of clients \cite{aono2017privacy,melis2019exploiting}.\\

\vspace{-0.3cm}
\subsection{3.2 Training challenges} 
\vspace{-0.05cm}
Training-related challenges encompass the issues related to high dimensional models, the overhead required during multiple training iterations, and heterogeneity of the models participating in the learning. Here, we will focus on high dimensionality, heterogeneous architectures, optimization of defense mechanisms, and challenges of non-i.i.d. datasets. 
 
\vspace{-0.05cm}
\subsection{(A) Curse of Dimensionality}
\vspace{-0.05cm}
Large models with huge dimensional parameter vectors are particularly susceptible to privacy, and security attacks \cite{gao2019privacy,chang2019cronus}. Most federated learning algorithms require overwriting the local model parameters with the global model, making them susceptible to poisoning and backdoor attacks. The adversary can make minuscule but detrimental changes in high-dimensional machine learning models without being detected. Thus, sharing the model parameters may not be an ideal design choice in federated learning as it opens all the model's internal state to inference attacks and maximizes the model's malleability by poisoning attacks. Understanding and determining the need for sharing model updates is essential to address these fundamental shortcomings of federated learning. Sharing less sensitive information or only sharing model predictions in a black-box manner can result in more robust privacy protection for federated learning \cite{gao2019privacy}.

\vspace{-0.05cm}
\subsection{(B) Heterogeneous Architectures}

\vspace{-0.05cm}
Sharing model updates are presently limited to homogeneous federated learning architectures. However, it would be compelling to study how to collaboratively extend federated learning to train models with heterogeneous architectures and investigate if state-of-the-art privacy preserving techniques are suited to such heterogeneous federated learning paradigms \cite{gao2019privacy,chang2019cronus}. 

\vspace{-0.05cm}
\subsection*{(C) Decentralized Federated Learning}
\vspace{-0.05cm}
Decentralized federated learning is a potential learning framework for collaboration among businesses that do not trust any third party as no centralized server is required in the system. In this criterion, each party could be voted in as a server in a round-robin manner. It would be interesting to examine if existing threats on server-based federated learning apply in this scenario \cite{yang2019federated}. 
Any adversarial participant can steal the training data from its neighbors if  we conduct decentralized training  in a "ring all reduce" manner \cite{lyu2019towards}.
It might open new attack surfaces as there is a possibility that the last party selected as the server is more likely to effectively poison the whole model if it chooses to insert backdoors. This scenario resembles server-based federated learning models, which were more vulnerable to backdoor attacks in later training rounds nearing convergence. \\
\vspace{-0.05cm}
\subsection*{(D) Optimization for Defense Mechanisms}
\vspace{-0.05cm}
Federated learning servers incur an extra computational cost when deploying defense mechanisms to identify an adversary attacking the system. In addition, disparate defense mechanisms may have different effectiveness against various attacks and incur a diverse cost. Therefore, it is crucial to study the optimization methods for deploying multiple defense mechanisms/ deterrence measures. Game-theory frameworks hold exceptional promise in addressing this challenge.
\vspace{-0.05cm}
\subsection*{(E) Challenges on non-i.i.d. datasets}
\vspace{-0.05cm}
Although many remedies have been recommended for handling non-i.i.d. data distributions in federated learning, many challenges remain open. Federated learning contains many hyperparameters, e.g.,  the number of local epochs, the total number of clients, and client dropout probability, which vary from algorithm to algorithm, making it hard to benchmark the actual non-i.i.d. performance of these algorithms. Additionally, though few real image datasets are proposed, a universal homogeneous and heterogeneous benchmark dataset has still not emerged in the field of federated learning \cite{hsu2020federated}. Synthetic non-i.i.d. data generated by arbitrary partitioning datasets may not effectively evaluate the performance of a method proposed for handling non-i.i.d. data. Though the vertical federated learning framework can be widely adopted in practical industry scenarios, only a handful of work has attempted to cope with the potential problems caused by non-i.i.d. distribution. The issue of overlapping data features, non-i.i.d. cases with both attribute and label skewness, and features with crowdsourcing skew deserve more attention.\\

Privacy protection is an essential purpose of federated learning. Still, several methods designed to address non-i.i.d. data, such as data sharing and knowledge distillation, inevitably increase the risk of privacy exposure. It is still unclear to what extent these methods harm data privacy, as there are no quantitative measures to identify the degree of privacy leakage. \\

There is an increasing demand for Automated Machine Learning (AutoML) \cite{cui2019fast}, and there are practical examples of self-learning within the field of Neural Architecture Search (NAS). However, only a limited amount of research on the influence of non-i.i.d. distribution on federated NAS has been reported \cite{he2020fednas,singh2020differentially}. Adversarial training was primarily developed for i.i.d. data and remains a challenging problem on how it can be adapted for non-i.i.d. settings.\\

As future directions, it would be worthwhile to have clearly defined quantitative criteria for measuring the degree of privacy leakage so that the maximum amount of shared data can be bounded. In order to compare federated learning algorithms fairly, the industry needs defined benchmark problems that reflect real word requirements and challenges along with standardized federated learning hyperparameter settings. Federated neural architecture search (FNAS)  is an emerging research direction, and handling non-i.i.d. problems in FNAS along with vertical federated learning is an interesting future direction \cite{zhao2018federated}.
\vspace{-0.1cm}
\section{4. Promising Research Directions}
\vspace{0.05cm}
 Federated learning finds excellent applications in almost every industry as it removes the barriers related to data sharing. Banking, financial services, healthcare, Internet-of-things (IoT), and natural language processing (NLP) applications related to next-word prediction and content suggestions represent promising areas to apply federated learning to increase data security and privacy.
 
\vspace{-0.1cm}
\subsection{4.1 Applications in Banking \& Financial Services}
One of the best uses of federated learning in finance is in the banking sector for example in \textbf{credit risk assessment} \cite{cheng2020federated, yang2019federated}. Typically banks use white-listing techniques to rule out the customers using their credit card reports from the central banks. Factors such as taxation and reputation could be prescribed for \textbf{risk management} by collaborating with other financial institutions and e-commerce companies. Federated learning could help build a risk assessment machine learning model to keep customer's information private among organizations. Banks  can leverage federated learning technologies for \textbf{credit risk management} in financial applications. Several banks could jointly generate a total credit score for a customer without sharing their data. With the development of research in federated learning, many companies or research teams can establish various tools oriented to federated learning-based research and  subsequent product development \cite{kawa2019credit}.\\

Financial institutions can train deep learning federated learning models on the server by sending encrypted model weights and bias coefficients back and forth \cite{yang2019federated}. Federated learning systems maintain client confidentiality relating to the portfolio components and have been used to optimize expense ratios and pricing for \textbf{portfolio management} in banking and financial services. These techniques allow managers, financial advisors, and robo-advisors to connect with other investment banks who can provide a fair purchase price during buying or selling a client's portfolio. Federated learning systems can potentially improve current efforts to curb unlawful financial activity like money laundering and fraud in coping with financial crimes as well as enhance regulatory compliance efforts such as to the GDPR. Federated learning techniques can improve this by enabling shared machine learning without sharing data.\\

Another avenue is the \textbf{open banking eco-system}. Open banking empowers individual customers to own their banking data with substantial potential benefits of customer experience, revenue, and the inclusion of more small and medium-sized players with innovative ideas and fine-grained service models \cite{long2020federated,brodsky2017data}. The most impressive aspect of federated learning is its ability to decompose model training into distributed nodes and a centralized server without collecting private data. This kind of disintegrated learning framework has great potential to protect user's privacy and sensitive data, and therefore, federated learning combines naturally with open banking data marketplaces. With federated learning,  it is foreseeable to have decentralized data ownership in the finance sector, thereby boosting a new ecosystem of data marketplaces and financial services. This just-in-time technology can learn intelligent models in a decentralized training manner. 
\vspace{-0.2cm}
\subsection {4.2 Applications in IoT and smart retail }
\vspace{-0.01cm}
The concept of edge computing as computing simple queries across distributed, low-powered devices has been investigated for over a decade in the topic of fog computing, computing at the edge, and sensor networks. Federated learning is an excellent fit for resource-constrained mobile devices, Internet-of-things (IoT), industrial sensor applications, smart retail, and other privacy-sensitive use cases.\\

Modern \textbf{IoT systems}, such as wearable devices, smart homes, autonomous vehicles, contain numerous sensors that collect and adapt to incoming data in real-time. For example, a fleet of independent sensors for self-driving vehicles may require up-to-date traffic or pedestrian behavior model to operate safely for traffic flow prediction techniques \cite{samarakoon2018federated}. The dynamic nature of the surroundings constrains existing autonomous driving decisions due to offline training. Building aggregate models among various organizations may be challenging due to the limited connectivity of each device and the private nature of the data. Federated learning methods can help train online models that efficiently adapt to changes in these systems while maintaining user privacy \cite{tan2020federated,liu2020privacy}.\\

Federated learning can also be an excellent application in \textbf{smart retail} as smart retail aims to use  machine learning technology to provide personalized services to customers based on customer data such as  user purchasing power and product characteristics for product recommendation and sales services \cite{zhao2020mobile}.

\subsection {4.3 Application in Healthcare}
Electronic Health Records (EHR) are considered the primary healthcare data sources for machine learning applications \cite{miotto2018deep}. Training machine learning models only using the limited data available in a single hospital might introduce bias in the predictions. Making the models more generalizable requires training with more data from a single hospital, which can only be realized by sharing data among organizations. It might not be feasible to share patient's electronic health records among hospitals considering the sensitive nature of data. Federated learning can be an excellent option for building a robust collaborative learning model and bringing together the research knowledge from different medical institutions \cite{min2019predictive}.

\subsection {4.4 Application in Natural Language Processing (NLP)}
 Machine learning and natural language processing have several facets: conversational dialogue systems, information extraction, structured prediction, clustering, language understanding, topic modeling, and ranking. NLP helps us better understand human language semantics. Building an NLP framework requires considerable amount of data to train highly accurate language models from multiple sources such as mobile phones, tablets, etc. However, privacy comes as a bottleneck for centralized language learning models as information from each edge device contains individual user information that needs to be protected. We can deploy federated learning methods to address these growing data risks,  data rights, privacy, and security \cite{garcia2020decentralizing}. It would be compelling to investigate the explainability of federated learning for NLP \cite{mathews2019explainable}, especially interpreting how NLP models work in data heterogeneity.

\vspace{-0.15cm}
\section{ 5. Conclusion and Future Work}
\vspace{0.05cm}
Federated learning is a new learning paradigm with a recent surge in popularity and helps train a high-quality shared global model with a central server from decentralized data scattered among several clients. As research in federated learning is still in the nascent stage, we believe that the issues presented in this paper are pivotal in shaping the developments in this area. We provided an overview of the existing system abstractions and building blocks for different federated learning systems. We have presented open-source tools to facilitate both the reproducibility of federated learning results and the dissemination of new solutions. \\

This article investigates the existing training and security challenges in federated learning and discusses the corresponding solutions. The proposed discussions can help build fully-fledged solutions for data privacy protection via federated learning. Applications related to finance, banking, healthcare, autonomous-driving, and the IoT ecosystem are promising candidate areas for future exploration for federated learning. Finally, we outlined a set of open problems that need to be addressed for federated learning to have a broader impact.

\newpage

\bibliography{aaai22.bib}

\begin{thebibliography}{58}
\providecommand{\natexlab}[1]{#1}

\bibitem[{Fed(AI Website)}]{Fed)}
 AI Website.
\newblock {Federated AI technology Enabler}.
\newblock \emph{https://www.fedai.org/cn/}.

\bibitem[{Git(Tensorflow)}]{Github}
 Tensorflow.
\newblock {Google: Tensorflow federated.}
\newblock \emph{https://www.tensorflow.org/federated.}

\bibitem[{Aledhari et~al.(2020)Aledhari, Razzak, Parizi, and
  Saeed}]{aledhari2020federated}
Aledhari, M.; Razzak, R.; Parizi, R.~M.; and Saeed, F. 2020.
\newblock Federated learning: A survey on enabling technologies, protocols, and
  applications.
\newblock \emph{IEEE Access}, 8: 140699--140725.

\bibitem[{Aono et~al.(2017)Aono, Hayashi, Wang, Moriai
  et~al.}]{aono2017privacy}
Aono, Y.; Hayashi, T.; Wang, L.; Moriai, S.; et~al. 2017.
\newblock Privacy-preserving deep learning via additively homomorphic
  encryption.
\newblock \emph{IEEE Transactions on Information Forensics and Security},
  13(5): 1333--1345.

\bibitem[{Bagdasaryan et~al.(2020)Bagdasaryan, Veit, Hua, Estrin, and
  Shmatikov}]{bagdasaryan2020backdoor}
Bagdasaryan, E.; Veit, A.; Hua, Y.; Estrin, D.; and Shmatikov, V. 2020.
\newblock How to backdoor federated learning.
\newblock In \emph{International Conference on Artificial Intelligence and
  Statistics}, 2938--2948. PMLR.

\bibitem[{Bhagoji et~al.(2019)Bhagoji, Chakraborty, Mittal, and
  Calo}]{bhagoji2019analyzing}
Bhagoji, A.~N.; Chakraborty, S.; Mittal, P.; and Calo, S. 2019.
\newblock Analyzing federated learning through an adversarial lens.
\newblock In \emph{International Conference on Machine Learning}, 634--643.
  PMLR.

\bibitem[{Biggio, Nelson, and Laskov(2012)}]{biggio2012poisoning}
Biggio, B.; Nelson, B.; and Laskov, P. 2012.
\newblock Poisoning attacks against support vector machines.
\newblock \emph{arXiv preprint arXiv:1206.6389}.

\bibitem[{Bonawitz et~al.(2017)Bonawitz, Ivanov, Kreuter, Marcedone, McMahan,
  Patel, Ramage, Segal, and Seth}]{bonawitz2017practical}
Bonawitz, K.; Ivanov, V.; Kreuter, B.; Marcedone, A.; McMahan, H.~B.; Patel,
  S.; Ramage, D.; Segal, A.; and Seth, K. 2017.
\newblock Practical secure aggregation for privacy-preserving machine learning.
\newblock In \emph{proceedings of the 2017 ACM SIGSAC Conference on Computer
  and Communications Security}, 1175--1191.

\bibitem[{Brodsky and Oakes(2017)}]{brodsky2017data}
Brodsky, L.; and Oakes, L. 2017.
\newblock Data sharing and open banking.
\newblock \emph{McKinsey \& Company}, 1097--1105.

\bibitem[{Chai et~al.(2020)Chai, Wang, Chen, and Yang}]{chai2020secure}
Chai, D.; Wang, L.; Chen, K.; and Yang, Q. 2020.
\newblock Secure federated matrix factorization.
\newblock \emph{IEEE Intelligent Systems}.

\bibitem[{Chang et~al.(2019)Chang, Shejwalkar, Shokri, and
  Houmansadr}]{chang2019cronus}
Chang, H.; Shejwalkar, V.; Shokri, R.; and Houmansadr, A. 2019.
\newblock Cronus: Robust and heterogeneous collaborative learning with
  black-box knowledge transfer.
\newblock \emph{arXiv preprint arXiv:1912.11279}.

\bibitem[{Chen et~al.(2017)Chen, Liu, Li, Lu, and Song}]{chen2017targeted}
Chen, X.; Liu, C.; Li, B.; Lu, K.; and Song, D. 2017.
\newblock Targeted backdoor attacks on deep learning systems using data
  poisoning.
\newblock \emph{arXiv preprint arXiv:1712.05526}.

\bibitem[{Cheng et~al.(2021)Cheng, Fan, Jin, Liu, Chen, Papadopoulos, and
  Yang}]{cheng2021secureboost}
Cheng, K.; Fan, T.; Jin, Y.; Liu, Y.; Chen, T.; Papadopoulos, D.; and Yang, Q.
  2021.
\newblock Secureboost: A lossless federated learning framework.
\newblock \emph{IEEE Intelligent Systems}.

\bibitem[{Cheng et~al.(2020)Cheng, Liu, Chen, and Yang}]{cheng2020federated}
Cheng, Y.; Liu, Y.; Chen, T.; and Yang, Q. 2020.
\newblock Federated learning for privacy-preserving AI.
\newblock \emph{Communications of the ACM}, 63(12): 33--36.

\bibitem[{Cui et~al.(2019)Cui, Chen, Li, Liu, Shen, and Jia}]{cui2019fast}
Cui, J.; Chen, P.; Li, R.; Liu, S.; Shen, X.; and Jia, J. 2019.
\newblock Fast and practical neural architecture search.
\newblock In \emph{Proceedings of the IEEE/CVF International Conference on
  Computer Vision}, 6509--6518.

\bibitem[{Dwork et~al.(2006)Dwork, Kenthapadi, McSherry, Mironov, and
  Naor}]{dwork2006our}
Dwork, C.; Kenthapadi, K.; McSherry, F.; Mironov, I.; and Naor, M. 2006.
\newblock Our data, ourselves: Privacy via distributed noise generation.
\newblock In \emph{Annual International Conference on the Theory and
  Applications of Cryptographic Techniques}, 486--503. Springer.

\bibitem[{Fontaine and Galand(2007)}]{fontaine2007survey}
Fontaine, C.; and Galand, F. 2007.
\newblock A survey of homomorphic encryption for nonspecialists.
\newblock \emph{EURASIP Journal on Information Security}, 2007: 1--10.

\bibitem[{Fung, Yoon, and Beschastnikh(2018)}]{fung2018mitigating}
Fung, C.; Yoon, C.~J.; and Beschastnikh, I. 2018.
\newblock Mitigating sybils in federated learning poisoning.
\newblock \emph{arXiv preprint arXiv:1808.04866}.

\bibitem[{Gao et~al.(2019)Gao, Liu, Huang, Ju, Yu, and Yang}]{gao2019privacy}
Gao, D.; Liu, Y.; Huang, A.; Ju, C.; Yu, H.; and Yang, Q. 2019.
\newblock Privacy-preserving heterogeneous federated transfer learning.
\newblock In \emph{2019 IEEE International Conference on Big Data (Big Data)},
  2552--2559. IEEE.

\bibitem[{Garcia~Bernal(2020)}]{garcia2020decentralizing}
Garcia~Bernal, D. 2020.
\newblock Decentralizing Large-Scale Natural Language Processing with Federated
  Learning.

\bibitem[{Gu, Dolan-Gavitt, and Garg(2017)}]{gu2017badnets}
Gu, T.; Dolan-Gavitt, B.; and Garg, S. 2017.
\newblock Badnets: Identifying vulnerabilities in the machine learning model
  supply chain.
\newblock \emph{arXiv preprint arXiv:1708.06733}.

\bibitem[{Hahn, Munir, and Mohanty(2018)}]{hahn2018security}
Hahn, D.~A.; Munir, A.; and Mohanty, S.~P. 2018.
\newblock Security and privacy issues in contemporary consumer electronics
  [energy and security].
\newblock \emph{IEEE Consumer Electronics Magazine}, 8(1): 95--99.

\bibitem[{Hardy et~al.(2017)Hardy, Henecka, Ivey-Law, Nock, Patrini, Smith, and
  Thorne}]{hardy2017private}
Hardy, S.; Henecka, W.; Ivey-Law, H.; Nock, R.; Patrini, G.; Smith, G.; and
  Thorne, B. 2017.
\newblock Private federated learning on vertically partitioned data via entity
  resolution and additively homomorphic encryption.
\newblock \emph{arXiv preprint arXiv:1711.10677}.

\bibitem[{He, Annavaram, and Avestimehr(2020)}]{he2020fednas}
He, C.; Annavaram, M.; and Avestimehr, S. 2020.
\newblock Fednas: Federated deep learning via neural architecture search.
\newblock \emph{arXiv e-prints}, arXiv--2004.

\bibitem[{Hsu, Qi, and Brown(2020)}]{hsu2020federated}
Hsu, T.-M.~H.; Qi, H.; and Brown, M. 2020.
\newblock Federated visual classification with real-world data distribution.
\newblock In \emph{Computer Vision--ECCV 2020: 16th European Conference,
  Glasgow, UK, August 23--28, 2020, Proceedings, Part X 16}, 76--92. Springer.

\bibitem[{Kawa et~al.(2019)Kawa, Punyani, Nayak, Karkera, and
  Jyotinagar}]{kawa2019credit}
Kawa, D.; Punyani, S.; Nayak, P.; Karkera, A.; and Jyotinagar, V. 2019.
\newblock Credit risk assessment from combined bank records using federated
  learning.
\newblock \emph{International Research Journal of Engineering and Technology
  (IRJET)}, 6(4): 1355--1358.

\bibitem[{Khan et~al.(2019)Khan, Aalsalem, Khan, and Arshad}]{khan2019data}
Khan, W.~Z.; Aalsalem, M.~Y.; Khan, M.~K.; and Arshad, Q. 2019.
\newblock Data and privacy: Getting consumers to trust products enabled by the
  Internet of Things.
\newblock \emph{IEEE Consumer Electronics Magazine}, 8(2): 35--38.

\bibitem[{Kone{\v{c}}n{\`y}, McMahan, and Ramage(2015)}]{konevcny2015federated}
Kone{\v{c}}n{\`y}, J.; McMahan, B.; and Ramage, D. 2015.
\newblock Federated optimization: Distributed optimization beyond the
  datacenter.
\newblock \emph{arXiv preprint arXiv:1511.03575}.

\bibitem[{Kone{\v{c}}n{\`y} et~al.(2016)Kone{\v{c}}n{\`y}, McMahan, Ramage, and
  Richt{\'a}rik}]{konevcny2016federated}
Kone{\v{c}}n{\`y}, J.; McMahan, H.~B.; Ramage, D.; and Richt{\'a}rik, P. 2016.
\newblock Federated optimization: Distributed machine learning for on-device
  intelligence.
\newblock \emph{arXiv preprint arXiv:1610.02527}.

\bibitem[{Leroy et~al.(2019)Leroy, Coucke, Lavril, Gisselbrecht, and
  Dureau}]{leroy2019federated}
Leroy, D.; Coucke, A.; Lavril, T.; Gisselbrecht, T.; and Dureau, J. 2019.
\newblock Federated learning for keyword spotting.
\newblock In \emph{ICASSP 2019-2019 IEEE International Conference on Acoustics,
  Speech and Signal Processing (ICASSP)}, 6341--6345. IEEE.

\bibitem[{Li et~al.(2019)Li, Wen, Wu, Hu, Wang, Li, Liu, and He}]{li2019survey}
Li, Q.; Wen, Z.; Wu, Z.; Hu, S.; Wang, N.; Li, Y.; Liu, X.; and He, B. 2019.
\newblock A survey on federated learning systems: vision, hype and reality for
  data privacy and protection.
\newblock \emph{arXiv preprint arXiv:1907.09693}.

\bibitem[{Li, Sharma, and Mohanty(2020)}]{li2020preserving}
Li, Z.; Sharma, V.; and Mohanty, S.~P. 2020.
\newblock Preserving data privacy via federated learning: Challenges and
  solutions.
\newblock \emph{IEEE Consumer Electronics Magazine}, 9(3): 8--16.

\bibitem[{Lim et~al.(2020)Lim, Luong, Hoang, Jiao, Liang, Yang, Niyato, and
  Miao}]{lim2020federated}
Lim, W. Y.~B.; Luong, N.~C.; Hoang, D.~T.; Jiao, Y.; Liang, Y.-C.; Yang, Q.;
  Niyato, D.; and Miao, C. 2020.
\newblock Federated learning in mobile edge networks: A comprehensive survey.
\newblock \emph{IEEE Communications Surveys \& Tutorials}, 22(3): 2031--2063.

\bibitem[{Lin et~al.(2020)Lin, Kong, Stich, and Jaggi}]{lin2020ensemble}
Lin, T.; Kong, L.; Stich, S.~U.; and Jaggi, M. 2020.
\newblock Ensemble distillation for robust model fusion in federated learning.
\newblock \emph{arXiv preprint arXiv:2006.07242}.

\bibitem[{Liu et~al.(2020{\natexlab{a}})Liu, James, Kang, Niyato, and
  Zhang}]{liu2020privacy}
Liu, Y.; James, J.; Kang, J.; Niyato, D.; and Zhang, S. 2020{\natexlab{a}}.
\newblock Privacy-preserving traffic flow prediction: A federated learning
  approach.
\newblock \emph{IEEE Internet of Things Journal}, 7(8): 7751--7763.

\bibitem[{Liu et~al.(2020{\natexlab{b}})Liu, Kang, Xing, Chen, and
  Yang}]{liu2020secure}
Liu, Y.; Kang, Y.; Xing, C.; Chen, T.; and Yang, Q. 2020{\natexlab{b}}.
\newblock A secure federated transfer learning framework.
\newblock \emph{IEEE Intelligent Systems}, 35(4): 70--82.

\bibitem[{Long et~al.(2020)Long, Tan, Jiang, and Zhang}]{long2020federated}
Long, G.; Tan, Y.; Jiang, J.; and Zhang, C. 2020.
\newblock Federated learning for open banking.
\newblock In \emph{Federated learning}, 240--254. Springer.

\bibitem[{Mathews(2019)}]{mathews2019explainable}
Mathews, S.~M. 2019.
\newblock Explainable artificial intelligence applications in NLP, biomedical,
  and malware classification: a literature review.
\newblock In \emph{Intelligent computing-proceedings of the computing
  conference}, 1269--1292. Springer.

\bibitem[{Melis et~al.(2019)Melis, Song, De~Cristofaro, and
  Shmatikov}]{melis2019exploiting}
Melis, L.; Song, C.; De~Cristofaro, E.; and Shmatikov, V. 2019.
\newblock Exploiting unintended feature leakage in collaborative learning.
\newblock In \emph{2019 IEEE Symposium on Security and Privacy (SP)}, 691--706.
  IEEE.

\bibitem[{Min, Yu, and Wang(2019)}]{min2019predictive}
Min, X.; Yu, B.; and Wang, F. 2019.
\newblock Predictive modeling of the hospital readmission risk from patient’s
  claims data using machine learning: a case study on COPD.
\newblock \emph{Scientific reports}, 9(1): 1--10.

\bibitem[{Miotto et~al.(2018)Miotto, Wang, Wang, Jiang, and
  Dudley}]{miotto2018deep}
Miotto, R.; Wang, F.; Wang, S.; Jiang, X.; and Dudley, J.~T. 2018.
\newblock Deep learning for healthcare: review, opportunities and challenges.
\newblock \emph{Briefings in bioinformatics}, 19(6): 1236--1246.

\bibitem[{Pan and Yang(2009)}]{pan2009survey}
Pan, S.~J.; and Yang, Q. 2009.
\newblock A survey on transfer learning.
\newblock \emph{IEEE Transactions on knowledge and data engineering}, 22(10):
  1345--1359.

\bibitem[{Ryffel et~al.(2018)Ryffel, Trask, Dahl, Wagner, Mancuso, Rueckert,
  and Passerat-Palmbach}]{ryffel2018generic}
Ryffel, T.; Trask, A.; Dahl, M.; Wagner, B.; Mancuso, J.; Rueckert, D.; and
  Passerat-Palmbach, J. 2018.
\newblock A generic framework for privacy preserving deep learning.
\newblock \emph{arXiv preprint arXiv:1811.04017}.

\bibitem[{Samarakoon et~al.(2018)Samarakoon, Bennis, Saad, and
  Debbah}]{samarakoon2018federated}
Samarakoon, S.; Bennis, M.; Saad, W.; and Debbah, M. 2018.
\newblock Federated learning for ultra-reliable low-latency V2V communications.
\newblock In \emph{2018 IEEE Global Communications Conference (GLOBECOM)},
  1--7. IEEE.

\bibitem[{Shafahi et~al.(2018)Shafahi, Huang, Najibi, Suciu, Studer, Dumitras,
  and Goldstein}]{shafahi2018poison}
Shafahi, A.; Huang, W.~R.; Najibi, M.; Suciu, O.; Studer, C.; Dumitras, T.; and
  Goldstein, T. 2018.
\newblock Poison frogs! targeted clean-label poisoning attacks on neural
  networks.
\newblock \emph{arXiv preprint arXiv:1804.00792}.

\bibitem[{Shokri et~al.(2017)Shokri, Stronati, Song, and
  Shmatikov}]{shokri2017membership}
Shokri, R.; Stronati, M.; Song, C.; and Shmatikov, V. 2017.
\newblock Membership inference attacks against machine learning models.
\newblock In \emph{2017 IEEE Symposium on Security and Privacy (SP)}, 3--18.
  IEEE.

\bibitem[{Singh et~al.(2020)Singh, Zhou, Yang, Ding, Lin, and
  Xie}]{singh2020differentially}
Singh, I.; Zhou, H.; Yang, K.; Ding, M.; Lin, B.; and Xie, P. 2020.
\newblock Differentially-private federated neural architecture search.
\newblock \emph{arXiv preprint arXiv:2006.10559}.

\bibitem[{Su and Xu(2018)}]{su2018securing}
Su, L.; and Xu, J. 2018.
\newblock Securing distributed machine learning in high dimensions.
\newblock \emph{arXiv preprint arXiv:1804.10140}.

\bibitem[{Tan et~al.(2020)Tan, Bremner, Le~Kernec, and
  Imran}]{tan2020federated}
Tan, K.; Bremner, D.; Le~Kernec, J.; and Imran, M. 2020.
\newblock Federated Machine Learning in Vehicular Networks: A summary of Recent
  Applications.
\newblock In \emph{2020 International Conference on UK-China Emerging
  Technologies (UCET)}, 1--4. IEEE.

\bibitem[{Truex et~al.(2019)Truex, Baracaldo, Anwar, Steinke, Ludwig, Zhang,
  and Zhou}]{truex2019hybrid}
Truex, S.; Baracaldo, N.; Anwar, A.; Steinke, T.; Ludwig, H.; Zhang, R.; and
  Zhou, Y. 2019.
\newblock A hybrid approach to privacy-preserving federated learning.
\newblock In \emph{Proceedings of the 12th ACM Workshop on Artificial
  Intelligence and Security}, 1--11.

\bibitem[{Xu et~al.(2021)Xu, Glicksberg, Su, Walker, Bian, and
  Wang}]{xu2021federated}
Xu, J.; Glicksberg, B.~S.; Su, C.; Walker, P.; Bian, J.; and Wang, F. 2021.
\newblock Federated learning for healthcare informatics.
\newblock \emph{Journal of Healthcare Informatics Research}, 5(1): 1--19.

\bibitem[{Yan, Guoliang, and Jianhua(2016)}]{yan2016survey}
Yan, Z.; Guoliang, L.; and Jianhua, F. 2016.
\newblock A survey on entity alignment of knowledge base.
\newblock \emph{Journal of Computer Research and Development}, 53(1): 165.

\bibitem[{Yang et~al.(2019)Yang, Liu, Chen, and Tong}]{yang2019federated}
Yang, Q.; Liu, Y.; Chen, T.; and Tong, Y. 2019.
\newblock Federated machine learning: Concept and applications.
\newblock \emph{ACM Transactions on Intelligent Systems and Technology (TIST)},
  10(2): 1--19.

\bibitem[{Yu et~al.(2020)Yu, Liu, Liu, Chen, Cong, Weng, Niyato, and
  Yang}]{yu2020sustainable}
Yu, H.; Liu, Z.; Liu, Y.; Chen, T.; Cong, M.; Weng, X.; Niyato, D.; and Yang,
  Q. 2020.
\newblock A sustainable incentive scheme for federated learning.
\newblock \emph{IEEE Intelligent Systems}, 35(4): 58--69.

\bibitem[{Zhang et~al.(2020)Zhang, Li, Xia, Wang, Yan, and
  Liu}]{zhang2020batchcrypt}
Zhang, C.; Li, S.; Xia, J.; Wang, W.; Yan, F.; and Liu, Y. 2020.
\newblock Batchcrypt: Efficient homomorphic encryption for cross-silo federated
  learning.
\newblock In \emph{2020 $\{$USENIX$\}$ Annual Technical Conference
  ($\{$USENIX$\}$$\{$ATC$\}$ 20)}, 493--506.

\bibitem[{Zhang et~al.(2021)Zhang, Xie, Bai, Yu, Li, and Gao}]{zhang2021survey}
Zhang, C.; Xie, Y.; Bai, H.; Yu, B.; Li, W.; and Gao, Y. 2021.
\newblock A survey on federated learning.
\newblock \emph{Knowledge-Based Systems}, 216: 106775.

\bibitem[{Zhao et~al.(2018)Zhao, Li, Lai, Suda, Civin, and
  Chandra}]{zhao2018federated}
Zhao, Y.; Li, M.; Lai, L.; Suda, N.; Civin, D.; and Chandra, V. 2018.
\newblock Federated learning with non-iid data.
\newblock \emph{arXiv preprint arXiv:1806.00582}.

\bibitem[{Zhao et~al.(2020)Zhao, Zhao, Jiang, Tan, and Niyato}]{zhao2020mobile}
Zhao, Y.; Zhao, J.; Jiang, L.; Tan, R.; and Niyato, D. 2020.
\newblock Mobile edge computing, blockchain and reputation-based crowdsourcing
  iot federated learning: A secure, decentralized and privacy-preserving
  system.

\end{thebibliography}

\end{document}